\documentclass[]{spie}  

\usepackage{amsmath,amsfonts,amssymb}
\usepackage{subcaption}
\usepackage{graphicx}
\usepackage{cite} 
\usepackage{epsfig}
\usepackage{amsmath}
\usepackage{nccmath}
\usepackage{amssymb}
\usepackage{mwe}
\usepackage{acro}
\usepackage{amssymb}
\usepackage{xcolor,colortbl}
\usepackage{tabularx}
\usepackage{relsize}
\usepackage{pifont}
\usepackage{booktabs} 
\usepackage{multirow}
\usepackage{multicol}
\usepackage{adjustbox}
\usepackage{float}
\usepackage{graphicx}
\usepackage{makecell}
\usepackage{tabu}
\usepackage[colorlinks=true, allcolors=blue]{hyperref}
\usepackage[capitalize]{cleveref}

\title{Persistence Image from 3D Medical Image: Superpixel and Optimized Gaussian Coefficient}
   
\author[a]{Yanfan Zhu}
\author[b]{Yash Singh}
\author[c]{Khaled Younis}
\author[a]{Shunxing Bao}
\author[d,a,e]{Yuankai Huo}

\affil[a]{Department of Electrical and Computer Engineering, Vanderbilt University, Nashville, TN, USA}

\affil[b]{Radiology, Mayo Clinic, Rochester, MN, USA}

\affil[c]{MedAiConsult, Cleveland, OH, USA}

\affil[d]{Department of Computer Science, Vanderbilt University, Nashville, TN, USA}

\affil[e]{Department of Pathology, Microbiology and Immunology, Vanderbilt University, Nashville Medical Center, TN, USA}

\authorinfo{Corresponding author: Yuankai Huo: E-mail: yuankai.huo@vanderbilt.edu}

\begin{document} 
\maketitle

\begin{abstract}
Topological data analysis (TDA) uncovers crucial properties of objects in medical imaging. Methods based on persistent homology have demonstrated their advantages in capturing topological features that traditional deep learning methods cannot detect in both radiology and pathology. However, previous research primarily focused on 2D image analysis, neglecting the comprehensive 3D context. In this paper, we propose an innovative 3D TDA approach that incorporates the concept of superpixels to transform 3D medical image features into point cloud data. By Utilizing Optimized Gaussian Coefficient, the proposed 3D TDA method, for the first time, efficiently generate holistic Persistence Images for 3D volumetric data. Our 3D TDA method exhibits superior performance on the MedMNist3D dataset when compared to other traditional methods, showcasing its potential effectiveness in modeling 3D persistent homology-based topological analysis when it comes to classification tasks. The source code is publicly available at \url{https://github.com/hrlblab/TopologicalDataAnalysis3D}.

\end{abstract}


\section{INTRODUCTION}
\label{sec:intro}  
In medical image research, particularly in the pathological analysis of 3D image collections such as MRI scans, these datasets are often directly input into deep learning networks designed to accept this image format. Alternatively, standard 2D convolutional layers are adapted into 3D convolutional layers to perform convolutions on three-dimensional data. A prominent example of this approach is the 3D ResNet\cite{hara2018can}, where 3D convolutional layers are employed to process volumetric data directly. Although this approach is convenient and widely adopted, the exponential increase in computational load and the large number of model parameters result in higher time costs and necessitate extensive data to prevent overfitting. This, in turn, increases the demands for hyperparameter tuning, model optimization, and data preprocessing. In their work, Singh et al. (2020)\cite{singh20203ddeeplearningmedical} underscore the challenges of applying deep learning to 3D medical images, noting that a large number of samples are essential for effective feature learning. However, the often limited sample sizes in 3D medical imaging pose difficulties in adapting to this scenario.

\begin{figure} [ht]
   \begin{center}
   \begin{tabular}{c} 
   \includegraphics[width=0.8\linewidth]{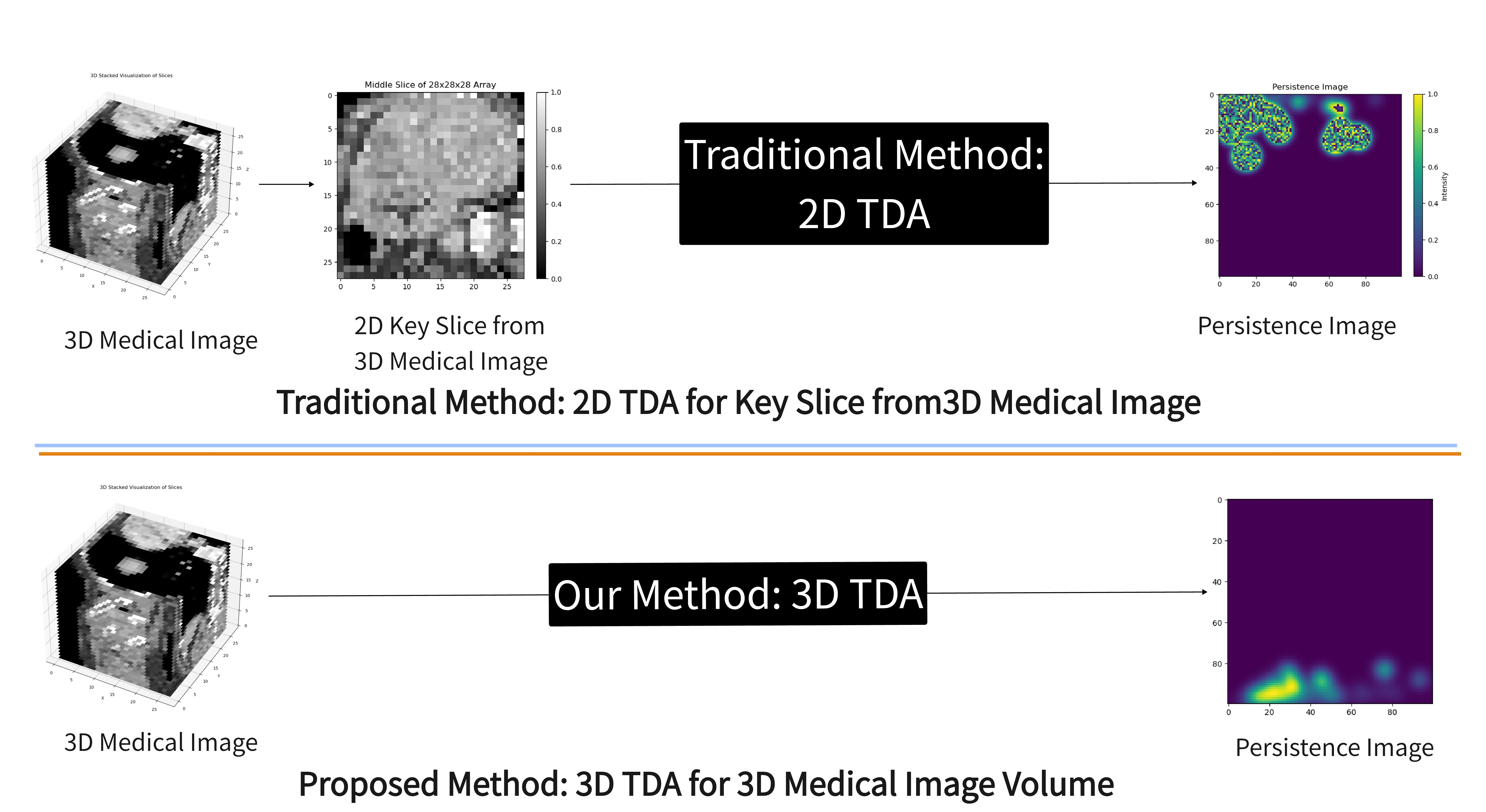}
   \end{tabular}
   \end{center}
   \caption[example] 
   { \label{fig:2d_tda_workflow} 
Compare Our Method to 2D TDA Method. Note that only one parameter is needed for the entire dataset to generate the Persistence Image, rather than a set of parameters, and our method results in less information loss.}
   \end{figure} 

In contrast, the integration of TDA and deep learning in medical imaging \cite{Singh2023} has garnered attention due to its distinctive approach of extracting features using persistent homology\cite{edelsbrunner2002topological}. In the study by Iqbal et al. (2021)\cite{iqbal2021classificationcovid19homologyctscan}, they constructed a feature point cloud (FPC) to reconstruct data, calculated Persistence Diagrams (PD)\cite{zomorodian2005computing} from two different filtered complexes, and generated a unique feature vector from CT scans of COVID-19 images. This work highlights the potential of using topological features extracted from different representations of 3D medical images to create distinct feature vectors.

Persistence Image (PI) is a method that converts the results of persistent homology in TDA into vector representations suitable for deep learning. PI effectively captures global topological features by weighting and Gaussian smoothing the PD. These features reflect the shape and structure of the data, which are often challenging for traditional methods to capture. Adams et al. (2017) extensively demonstrated the stability of PI in both general and Gaussian distributions\cite{Adams2017}. PI discretizes and Gaussian smooths PDs into fixed-size images (matrices), making them compatible with deep learning networks. The resulting feature vectors can be directly input into standard convolutional neural networks for training and prediction. Therefore, using intensity plots to expand 2D medical image slices into point clouds and then applying TDA to obtain PIs for training has become a viable method, as illustrated in Figure \ref{fig:2d_tda_workflow}. This approach effectively leverages the advantages of TDA to explore the topological correlations within the data. However, since slices within a 3D medical image are inherently related in a real-world context, learning from a single slice in isolation from a 3D medical image becomes impractical for real-world applications. Therefore, we need to learn from the entire 3D dataset. Directly computing persistent homology on the expanded dimensions of 3D medical images is computationally expensive, and current hardware does not support acceleration for this computation. Additionally, ensuring consistency in the discretization and Gaussian smoothing in PI is also a critical issue.

Therefore, this study combines the superpixel method proposed by Gautier et al. (2019)\cite{lggit} for unsupervised image segmentation using persistent homology theory and optimizes the extraction of 3D medical image features, extending the extracted features to higher-dimensional space. Through mathematical definitions, pixel-level constraint optimized Gaussian coefficients are proposed to determine the impact of Gaussian smoothing at the pixel level during the computation and vectorization of persistent homology into PI. Suitable homology groups and optimized Gaussian coefficients were determined through this research. The main contributions of this work are as follows:

\begin{itemize}
\item A superpixel method is provided, reducing the computational cost of constructing simplicial complexes and enabling the use of TDA to extract features from entire 3D medical images. This addresses the issue of computing persistent homology on large datasets.
\item An optimized Gaussian coefficient is proposed to solve the problem of inconsistent Gaussian sigma parameters required for generating PI plots from different datasets. This allows for better standardization of features during the model training process.
\end{itemize}

This method was compared to others that introduce 3D medical images as input in different multi-organ classification tasks. The results demonstrate superior performance compared to other methods, showcasing its ability to capture more features from 3D medical images. To the best of our knowledge, the method presented in this paper is the first solution to apply TDA to the entire 3D medical image domain. The source code has been made publicly available at \url{https://github.com/hrlblab/TopologicalDataAnalysis3D}.

\section{METHOD}
\subsection{Point Cloud Converter Based on Superpixels}
TDA focuses on the overarching shape and structure of data, revealing insights that are often obscured in the raw data alone. The core concept of TDA is the use of persistent homology, a tool that captures and analyzes topological features of a space at various spatial resolutions. Persistent homology examines the evolution of topological features across different scales within a parameterized sequence of subspaces, known as a filtration. A filtration of a topological space \(X\) is a nested sequence of subspaces:
\[
\emptyset = X_0 \subseteq X_1 \subseteq \cdots \subseteq X_n = X
\]
where each \(X_i\) represents a subspace at a specific parameter level \(i\), capturing data structures at varying resolutions.
Homology groups are computed at each level of the filtration to identify fundamental topological structures—such as connected components, holes, and voids. Persistent homology is particularly adept at tracking the continuity and changes of these features as they evolve through the filtration levels. Each topological feature is characterized by \(birth\) and \(death\) parameter, represent the filtration level at which the feature first appears and disappears. The longevity of a feature, defined by \(death - birth\), indicates its persistence across the filtration, signifying its potential importance in the underlying data structure. 
Mathematically, the structure of persistent homology is formalized through persistence modules, which are sequences of vector spaces connected by linear maps:
\[
H_k(X_0) \rightarrow H_k(X_1) \rightarrow \cdots \rightarrow H_k(X_n)
\]
where \(H_k(X_i)\) denotes the \(k\)-th homology group of the subspace \(X_i\). The mappings are induced by inclusions \(X_i \subseteq X_{i+1}\), emphasizing the continuation of homological features.
   \begin{figure} [t]
   \begin{center}
   \begin{tabular}{c} 
   \includegraphics[width=0.8\linewidth]{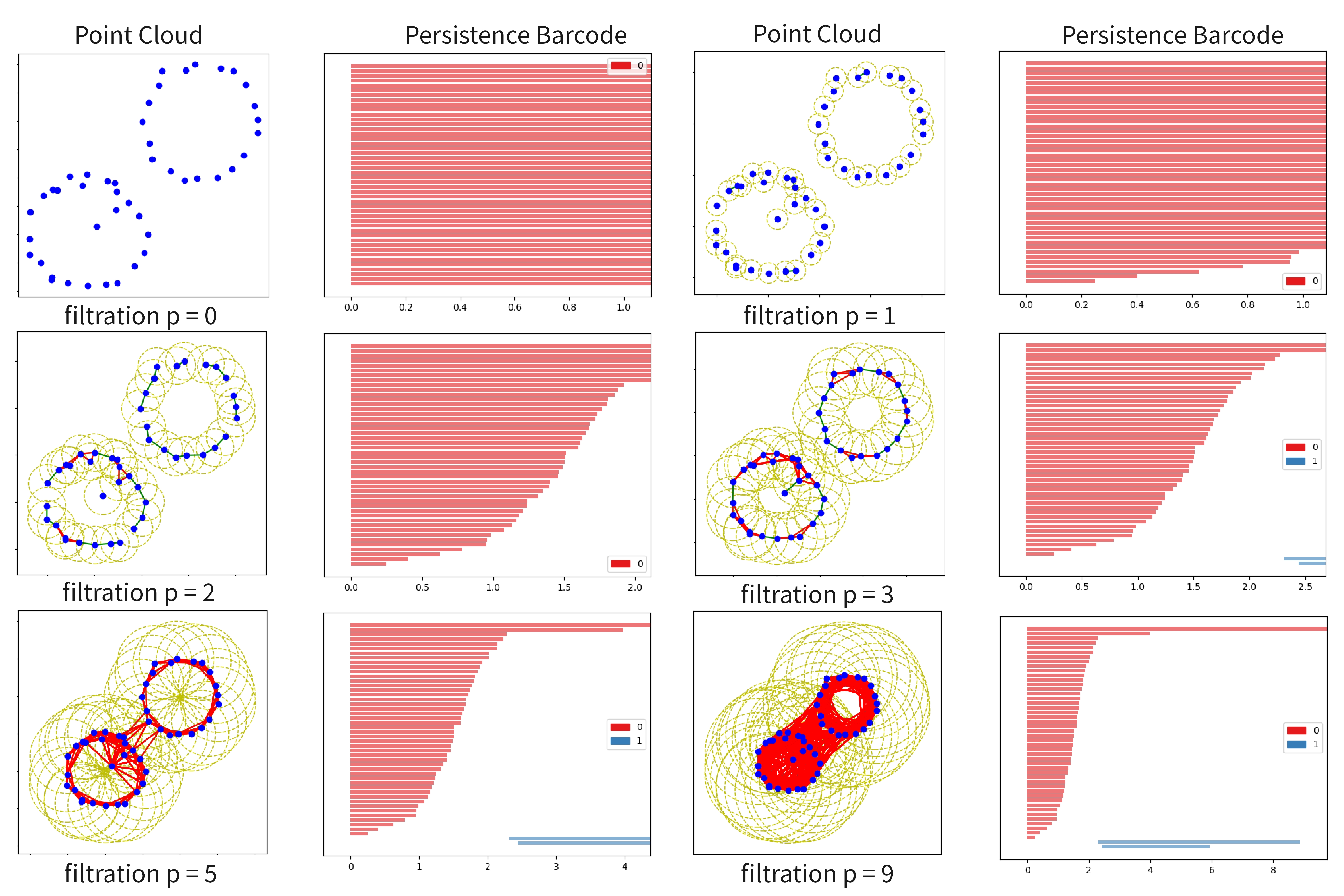}
   \end{tabular}
   \end{center}
   \caption[example] 
   { \label{fig:filtration} 
Persistent Homology Filtration Process. Each row in the barcode represents the birth and death of a new topological feature at a certain filtration value. Red indicates the 0-dimensional homology group, which represents connected components, while blue represents the 1-dimensional homology group, which captures loops or cycles within the data.}
   \end{figure} 

Figure \ref{fig:filtration} clearly shows the process of filtration while points in a point cloud reaching others and generating a Persistence Barcode which indicates the quantization metrics as birth and death of all points. Note that the red lines connected in the unions of balls represent higher dimension homology groups associated to the blue bars in the barcodes. To perform persistent homology on the inherent topological structure of a 3D medical image, the 3D medical image needs to be converted. However, calculating high-dimensional complexes requires huge computational resources, and the current hardware architecture does not support acceleration for this computation. Therefore, a feature point extraction method is employed to calculate the step size and density in the three-dimensional space using the estimated number of feature points (i.e., superpixels). These feature points are uniformly distributed throughout the 3D space, and Gaussian filtering is applied to ensure that the extracted superpixels accurately represent the features in the original 3D medical image.
\subsection{Optimized Gaussian Coefficient}
However, directly using PDs in deep learning model training is challenging due to their non-vector nature. PIs address this issue by mapping points from PDs onto a discretized grid and applying Gaussian kernel functions to generate a smooth and continuous representation. This conversion not only preserves the essential topological information but also creates a structured and stable feature set for further analysis\cite{chazal2021introduction}. The transformation from these non-vector features to a PI is defined as follows:

\begin{definition}
\label{def:pi}
A \textbf{Persistence Image (PI)} is a finite-dimensional vector representation of a persistence diagram $\mathcal{D}$ obtained through the following steps:
\begin{enumerate}
    \item \textbf{Weighting Function}: Define a weighting function $\rho: \mathbb{R}^2 \to \mathbb{R}$ to assign importance to each point $(b, d)$ in the PD:
    \begin{equation}
    \label{weightingfunction}
        \rho(b, d) = \exp \left( -\frac{(d - b)^2}{2\sigma^2} \right),
    \end{equation}
    
    where $d > b$. This weight corresponds to the persistence of the feature.
    
    \item \textbf{Gaussian Distribution}: Represent each point $(b, d) \in \mathcal{D}$ by a Gaussian distribution centered at $(b, d)$:
    \begin{equation}
    \label{gaussian distribution}
        \varphi_{\sigma}(x, y; b, d) = \frac{1}{2\pi\sigma^2} \exp \left( -\frac{(x - b)^2 + (y - d)^2}{2\sigma^2} \right)
    \end{equation}
    where $\sigma$ is the standard deviation controlling the spread of the Gaussian.
    
    \item \textbf{Discretization}: Discretize the plane $\mathbb{R}^2$ into a grid of pixels. For each pixel $(i, j)$ with center $(x_i, y_j)$, compute the value by summing the contributions of all Gaussians weighted by $\rho$:
    \begin{equation}
    \label{discretization}
        \mathrm{PI}(i, j) = \sum_{(b, d) \in \mathcal{D}} \rho(b, d) \cdot \varphi_{\sigma}(x_i, y_j; b, d).
    \end{equation}
    
    \item \textbf{Normalization}: To ensure numerical stability and comparability, normalize the resulting PI.
\end{enumerate}
\end{definition}

In the \textbf{Definition \ref{def:pi}}, the weight $\rho(b, d)$ emphasizes features that persist over a longer range. The choice of $\sigma$ influences the smoothness of the PI. A smaller $\sigma$ results in less smoothing and retains more details, whereas a larger $\sigma$ results in more smoothing and reduces noise. The resolution of the grid affects the level of detail in the PI. Based on the properties of $\sigma$, which affect the performance of the PI image during the training process, we explore its impact on pixel values after discretization to establish an adjustable coefficient.

We assume that only one persistent homology group of a certain dimension exists in this PI image, with a resolution of $k$, of which the range of birth is $m$ to $M$, and the range of death is $n$ to $N$. The difference between individual pixel values is $\Delta x = \frac{M - m}{k}$ and $\Delta y = \frac{N - n}{k}$. We set a ratio between two adjacent pixels as $\epsilon$, which is defined as:

\begin{equation}
\epsilon = \frac{\exp\left(-\frac{(x+\Delta x -b)^2+(y+\Delta y-d)^2}{2\sigma^2}\right)}{\exp\left(-\frac{(x -b)^2+(y-d)^2}{2\sigma^2}\right)},
\end{equation}

where the \textbf{Equation\ref{discretization}} reaches its maximum when $x=b, y=d$. Thus,

\begin{equation}
\label{optimizedsigma}
\sigma = \sqrt{\frac{-(\Delta x)^2 - (\Delta y)^2}{2\ln(\epsilon)}},
\end{equation}

where $\epsilon$ is a coefficient in the range $(0,1]$. The value of \(\sigma\) established in this manner is only influenced by \(\epsilon\). This parameter \(\epsilon\), besides having clear practical significance in the discretized PI image, ensures that the \(\sigma\) value determined through this parameter adjustment will not vary with changes in the resolution requirements of the original data in practical applications but will differ from different requirements of PI.

In this experimental method, the dimensions and pixels are first calculated, and then a suitable step size is selected based on the target superpixel count. After generating grid coordinates and applying Gaussian filtering, the image data is converted into four-dimensional point cloud data, comprising \(z\), \(y\), \(x\), and pixel intensity. The GUDHI library was utilized for topological data analysis \cite{gudhi}, to obtain alpha complexes from the generated simplex tree and generate the PI for vectorization. By adjusting the value of \(epsilon\) \textbf{Equation} \ref{optimizedsigma}, the Gaussian kernel in the PI is determined.
\begin{figure} [t]
\begin{center}
\begin{tabular}{c} 
\includegraphics[width=0.95\linewidth]{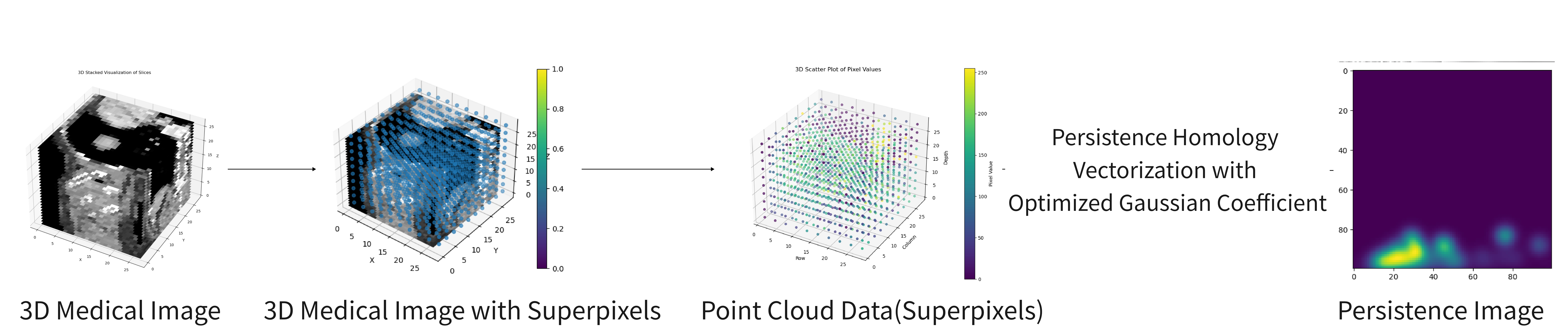}
\end{tabular}
\end{center}
\caption[example] 
{ \label{fig:preprocess} The dataset preprocessing workflow transforms a single 3D medical image into a PI.
}
\end{figure} 

Figure \ref{fig:preprocess} illustrates the transformation of a single 3D medical image into a PI. The transformed 2D array, with four dimensions, is represented in a three-dimensional space. In this representation, in addition to \(z\), \(y\) and \(x\), the original pixel intensity value is indicated by color intensity. Throughout the data preprocessing progress, the topological information in the image features is mapped onto a grayscale image with a specified resolution. After vectorization, this not only facilitates subsequent deep learning training but also reduces training costs. The grayscale PI, composed of this topological information, is used as the input for training the model.

\section{DATA AND EXPERIMENTAL DESIGN}
\label{sec:experiment}
\subsection{Data}
For our experimental evaluation, the MedMNist V2 3D Medical Image Dataset\cite{medmnistv2}, specifically designed for medical image analysis, is well-suited to our needs. The dataset includes various 3D medical images such as abdominal CT scans, chest CT scans, and brain MRA scans. It covers multiple classification tasks with different numbers of classes, ranging from 2 to 11. The total number of samples for each dataset varies, with splits between training, validation, and test sets. For example, the OrganMNIST3D dataset has 1,742 samples divided into 971 for training, 161 for validation, and 610 for testing. Similarly, other datasets like NoduleMNIST3D, AdrenalMNIST3D, FractureMNIST3D, VesselMNIST3D, and SynapseMNIST3D have their own specific distributions of samples across training, validation, and test sets, making them comprehensive and diverse for evaluating different medical imaging analysis tasks.

\subsection{Experimental Design}
600 superpixels were selected and our proposed method was applied to the Organ3D dataset, testing different persistent homology groups to obtain various PI images. These images were then used to train a standard ResNet18 \cite{he2015deepresiduallearningimage} to explore the correlation between the topological structure of the transformed superpixels and the original data. Additionally, we tested the training performance of PI images obtained with the optimized Gaussian coefficient ranging from 0.85 to 0.97. Subsequently, the method was applied to the four mentioned datasets, including a 2D TDA method, which transforms a single slice (the mid slice from the inferior axis was chosen) from the data into a 4D space and applies TDA to obtain PI. The results were compared with the benchmarking performance obtained using different methods, thereby comprehensively exploring the advantages and disadvantages of this method.

\section{RESULTS}

The performance of different homology groups and optimized Gaussian coefficients were compared. The 2D homology group stands out with a best performance of 0.924 and a worst of 0.915. Other homology groups showed consistent but lower performance. For Gaussian coefficients, performance gradually improved from 0.85 to 0.93. A jump occurred at 0.95, reaching 0.9179, indicating an optimal point. However, performance dropped to 0.4770 at 0.97.

\begin{table}[ht]
\caption{Performance comparison across different datasets and methods}
\label{tab:final}
\centering
\begin{tabular}{ p{4cm} m{1.5cm}<{\centering} m{1.5cm}<{\centering} m{1.5cm}<{\centering} m{1.5cm}<{\centering} m{1.5cm}<{\centering} m{1.5cm}<{\centering} }
\hline
\rule[-1ex]{0pt}{3.5ex}
 & \textbf{Organ} & \textbf{Adrenal} & \textbf{Nodule} & \textbf{Fracture} & \textbf{Vessel} & \textbf{Synapse} \\
 \hline
\rule[-1ex]{0pt}{3.5ex}
\textbf{2D-TDA} & 0.143 & N/A & 0.809 & 0.383 & N/A & 0.724 \\
\hline
\rule[-1ex]{0pt}{3.5ex}
\textbf{2.5D ResNet-18} & 0.762 & 0.681 & \textcolor{blue}{0.841} & \textcolor{blue}{0.451} & 0.814 & \textcolor{red}{0.801} \\
\hline
\rule[-1ex]{0pt}{3.5ex}
\textbf{3D Resnet-18} & 0.822 & 0.664 & \textcolor{red}{0.845} & 0.412 & \textcolor{red}{0.939} & 0.732 \\
\hline
\rule[-1ex]{0pt}{3.5ex}
\textbf{ACS ResNet-18} & \textcolor{blue}{0.911} & \textcolor{blue}{0.761} & 0.825 & 0.445 & 0.738 & 0.698 \\
\hline
\rule[-1ex]{0pt}{3.5ex}
\textbf{3D TDA (Ours)} & \textcolor{red}{0.931} & \textcolor{red}{0.768} & 0.809 & \textcolor{red}{0.512} & \textcolor{blue}{0.903} & \textcolor{blue}{0.735} \\
\hline
\end{tabular}
\end{table}

Based on the results we achieved, we selected the 2D homology group for representing PIs and set the optimized Gaussian coefficient to 0.95. We then conducted the training process on six different datasets, which include both binary-class and multi-class tasks. \textbf{Table} \ref{tab:final} provides a comparison of the results from traditional TDA method, our proposed method and other methods provided by MedMNist\cite{medmnist2023experiments}, including various versions of ResNet (with TDA-2D input, 2.5D, 3D, and ACS configurations). The best result for each dataset is highlighted in red, while the second best is highlighted in blue. Noticed that when using 2D TDA, we observed that it could not effectively learn the features of the Organ3D dataset. For the Adrenal3D and Vessel3D datasets, we were unable to generate the corresponding PI plots. Our proposed method demonstrated better performance compared to other methods on the majority of datasets, achieving the highest accuracy on Organ3D, Adrenal3D, and Fracture3D, and the second-highest accuracy on Vessel3D and Synapse3D. Especially on the Organ3D dataset, which has 11 classes, our method achieved a higher accuracy than the second-best result.

\section{DISCUSSION}

\label{sec:conclusion}        
The results from our extensive experiments across various 3D medical image datasets demonstrate that the combination of TDA and deep learning, specifically using PIs, offers a robust and competitive method for medical image analysis. Our proposed method achieved the highest accuracy on several datasets, indicating its effectiveness in capturing and leveraging the topological features inherent in 3D medical images.

The performance of 2D homology group which is superior than others could be attributed to the nature of the data and the specific features that the 2D homology group is designed to detect. In the context of medical imaging, this group might be more effective at identifying and preserving crucial structural information such as cavities or voids, which are important in medical diagnostics.

2D TDA method in our experimental result shows instability of unsuccessfully generating topology features which is largely due to the need for separate analyses to select an appropriate slice for each dataset. The method heavily relies on domain-specific knowledge of the dataset and often loses feature representations due to voxel spacing, making it difficult to serve as a standardized approach applicable to all datasets.

Our experiments with different optimized Gaussian coefficients revealed that a coefficient of 0.95 enhances model performance, striking an optimal balance between noise reduction and feature retention. This optimization process is crucial for ensuring that the PIs of 2D homology group generated are both informative and stable for deep learning applications. When compared to other methods, including various ResNet configurations, our method consistently provided competitive results. This highlights the advantage of integrating TDA with deep learning to better utilize the structural and topological information in the data.

However, it's important to note that while our method showed superior performance in several cases, other methods performed better on specific datasets, such as Nodule3D and Synapse3D. This indicates that the choice of method should be tailored to the specific characteristics of the dataset. In conclusion, our research demonstrates that performing superpixel feature extraction on entire 3D medical images and using an optimized Gaussian coefficient enhances the analysis and interpretation of 3D medical images compared to other methods. Future work could focus on improving the feature extraction process and exploring the use of hybrid approaches to leverage the strengths of various methods for even better performance.

\acknowledgments
This research was supported by NIH R01DK135597(Huo), DoD HT9425-23-1-0003(HCY), NIH NIDDK DK56942(ABF). This work was also supported by Vanderbilt Seed Success Grant, Vanderbilt Discovery Grant, and VISE Seed Grant. This project was supported by The Leona M. and Harry B. Helmsley Charitable Trust grant G-1903-03793 and G-2103-05128. This research was also supported by NIH grants R01EB033385, R01DK132338, REB017230, R01MH125931, and NSF 2040462. We extend gratitude to NVIDIA for their support by means of the NVIDIA hardware grant. This works was also supported by NSF NAIRR Pilot Award NAIRR240055.

\bibliography{report} 

\begin{thebibliography}{10}

\bibitem{hara2018can}
Hara, K., Kataoka, H., and Satoh, Y., ``Can spatiotemporal 3d cnns retrace the history of 2d cnns and imagenet?,'' in [{\em Proceedings of the IEEE Conference on Computer Vision and Pattern Recognition (CVPR)}{\nolinebreak\hspace{0.1em}]},   6546--6555 (2018).

\bibitem{singh20203ddeeplearningmedical}
Singh, S.~P., Wang, L., Gupta, S., Goli, H., Padmanabhan, P., and Gulyás, B., ``3d deep learning on medical images: A review,'' (2020).

\bibitem{Singh2023}
Singh, Y., Farrelly, C.~M., Hathaway, Q.~A., Leiner, T., Jagtap, J., Carlsson, G.~E., and Erickson, B.~J., ``Topological data analysis in medical imaging: current state of the art,'' {\em Insights into Imaging}~{\bf 14}(1),  58 (2023).

\bibitem{edelsbrunner2002topological}
Edelsbrunner, H., Letscher, D., and Zomorodian, A., ``Topological persistence and simplification,'' {\em Discrete \& Computational Geometry}~{\bf 28}(4),  511--533 (2002).

\bibitem{iqbal2021classificationcovid19homologyctscan}
Iqbal, S., Ahmed, H.~F., Qaiser, T., Qureshi, M.~I., and Rajpoot, N., ``Classification of covid-19 via homology of ct-scan,'' (2021).

\bibitem{zomorodian2005computing}
Zomorodian, A. and Carlsson, G., ``Computing persistent homology,'' {\em Discrete \& Computational Geometry}~{\bf 33}(2),  249--274 (2005).

\bibitem{Adams2017}
Adams, H., Emerson, T., Kirby, M., Neville, R., Peterson, C., Shipman, P., Chepushtanova, S., Hanson, E., Motta, F., and Ziegelmeier, L., ``Persistence images: A stable vector representation of persistent homology,'' {\em Journal of Machine Learning Research}~{\bf 18}(8),  1--35 (2017).

\bibitem{lggit}
Gautier, L. \url{https://github.com/salimandre/unsupervised-image-segmentation-persistent-homology} (Jan. 2019).

\bibitem{chazal2021introduction}
Chazal, F. and Michel, B., ``An introduction to topological data analysis: fundamental and practical aspects for data scientists,'' (2021).

\bibitem{gudhi}
Glisse, M., Bauer, U., Tierny, J., Cohen-Steiner, D., Boissonnat, J.-D., Edelsbrunner, H., Oudot, S., Wintraecken, M., Kerber, M., Morozov, D., Dlotko, A., Adams, H., and Skraba, P., ``Gudhi: Simplicial complexes and persistent homology.'' \url{http://gudhi.gforge.inria.fr/} (2014--).
\newblock Version 3.3.0.

\bibitem{medmnistv2}
Yang, J., Shi, R., Wei, D., Liu, Z., Zhao, L., Ke, B., Pfister, H., and Ni, B., ``Medmnist v2-a large-scale lightweight benchmark for 2d and 3d biomedical image classification,'' {\em Scientific Data}~{\bf 10}(1),  41 (2023).

\bibitem{he2015deepresiduallearningimage}
He, K., Zhang, X., Ren, S., and Sun, J., ``Deep residual learning for image recognition,'' (2015).

\bibitem{medmnist2023experiments}
MedMNIST, ``Medmnist experiments.'' \url{https://github.com/MedMNIST/experiments} (2023).

\end{thebibliography}
\bibliographystyle{spiebib} 

\end{document}